\documentclass[letterpaper]{article} 
\usepackage[draft]{aaai2026}  
\usepackage{times}  
\usepackage{helvet}  
\usepackage{courier}  
\usepackage[hyphens]{url}  
\usepackage{graphicx} 
\usepackage{natbib}  
\usepackage{caption} 
\usepackage{amsmath}
\usepackage{multirow}

\PassOptionsToPackage{table}{xcolor}
\usepackage{xcolor}
\usepackage{amsfonts} 
\usepackage{mathtools}
\usepackage[ruled,vlined,linesnumbered]{algorithm2e}
\SetKwComment{Comment}{$\triangleright$\ }{}
\SetKwInput{KwInput}{Input}
\SetKwInput{KwOutput}{Output}

\mathtoolsset{showonlyrefs=true}

\usepackage{booktabs}
\usepackage{newfloat}
\title{Generative AI Against Poaching: Latent Composite Flow Matching for Wildlife Conservation}

\author{
    Lingkai Kong\textsuperscript{\rm 1}\equalcontrib,
    Haichuan Wang\textsuperscript{\rm 1}\equalcontrib,
    Charles A. Emogor\textsuperscript{\rm 1,2},
    Vincent B\"orsch-Supan\textsuperscript{\rm 1},
    Lily Xu\textsuperscript{\rm 3},\\
    Milind Tambe\textsuperscript{\rm 1}
}

\affiliations {
    \textsuperscript{\rm 1}School of Engineering and Applied Sciences, Harvard University \\
    \textsuperscript{\rm 2}Department of Computer Science and Technology, University of Cambridge \\
    \textsuperscript{\rm 3}Department of Industrial Engineering and Operations Research, Columbia University 
}

\usepackage{xspace}

\newcommand{\ours}{\textsc{WildFlow}\xspace}

\newcommand{\std}[1]{\scriptsize
{$\pm$#1}}

\begin{document}

\maketitle

\begin{abstract}
Poaching poses significant threats to wildlife and biodiversity. A valuable step in reducing poaching is to forecast poacher behavior, which can inform patrol planning and other conservation interventions. Existing poaching prediction methods based on linear models or decision trees lack the expressivity to capture complex, nonlinear spatiotemporal patterns. Recent advances in generative modeling, particularly flow matching, offer a more flexible alternative. However, training such models on real-world poaching data faces two central obstacles: imperfect detection of poaching events and limited data. To address imperfect detection, we integrate flow matching with an occupancy-based detection model and train the flow in latent space to infer the underlying occupancy state. To mitigate data scarcity, we adopt a composite flow initialized from a linear-model prediction rather than random noise which is the standard in diffusion models, injecting prior knowledge and improving generalization. Evaluations on datasets from two national parks in Uganda show consistent gains in predictive accuracy.
\end{abstract}


\section{Introduction}

Poaching is a major, direct driver of wildlife loss. First, the toll on elephants is stark: in Africa, about 100{,}000 were illegally killed in 2010--2012 for ivory\citep{wittemyer2014illegal}. Beyond elephants, pangolins are targeted for their scales (and meat), with about 8.5 million taken for illegal trade in West and Central Africa from 2014--2021\citep{Oxford2023OperationPangolin,UNODC2020Pangolin}. Predators are not spared: in one monitored southern African population (2011--2018), poaching for body parts (including bones) caused 35\% of recorded human-caused lion deaths~\citep{everatt2019evidence}. These facts show the need to direct limited patrol resources to the places where they can best stop poaching, in line with global reports on wildlife trafficking \cite{UNODC2024WWCR}.

Given the scale of these losses, an effective way to strengthen protection is to forecast poacher behavior and use these predictions to guide patrol planning. Machine learning models have been applied to estimate poaching risk and support data-driven patrol allocation. However, existing approaches typically rely on classical models—such as linear models~\citep{xu2021robust} and decision trees~\citep{thiault2020predicting, kar2017cloudy}—which are unable to capture the complex spatial pattens and the strategic adaptations of poachers in response to patrols. These models often make simplifying assumptions about spatial independence, which can lead to inaccurate risk forecasts and, in turn, less effective patrol deployment.

\begin{figure}[t]
    \centering
\includegraphics[width=0.75\linewidth]{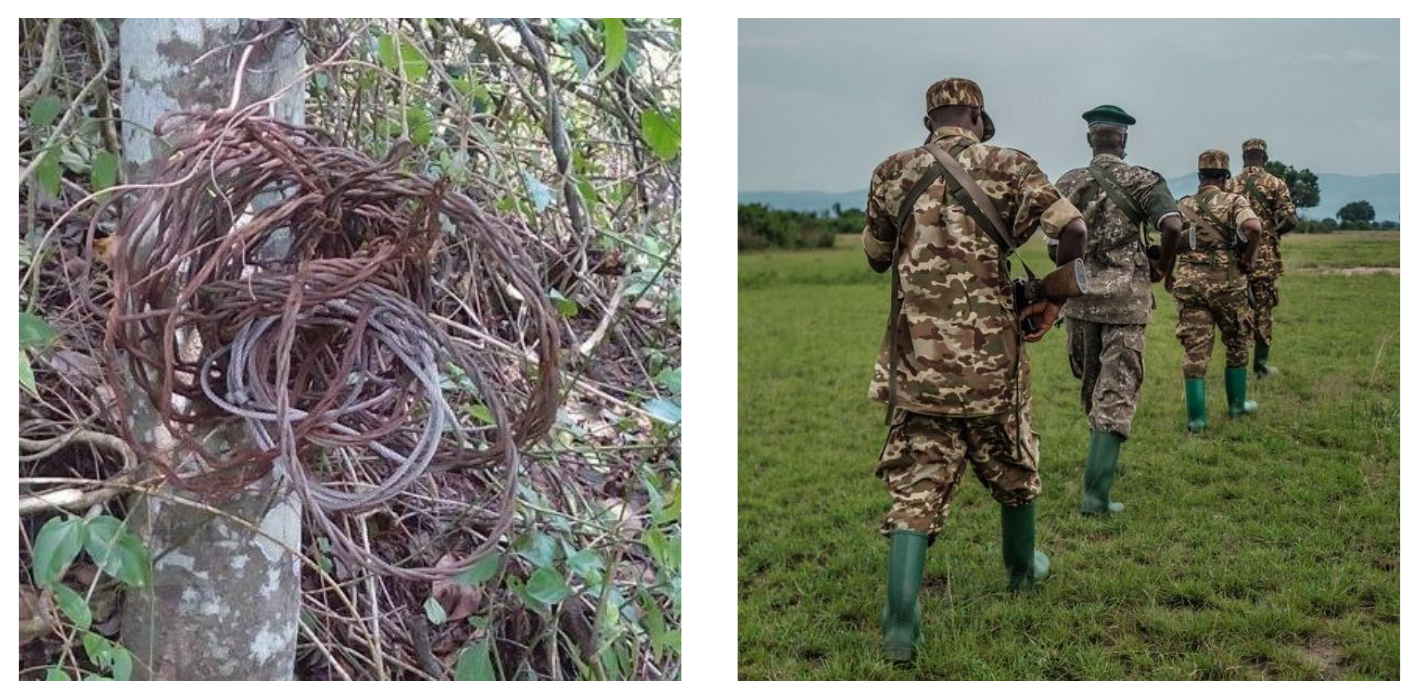}
    \caption{Well-hidden snares and rangers conducting a patrol to locate them. Photos: Uganda Wildlife Authority}
    \label{fig:snares}
\end{figure}

Recent advances in generative AI, particularly flow matching~\citep{lipman2023flow} and diffusion models~\citep{ho2020denoising}, offer powerful distribution-fitting capabilities that can model complex, high-dimensional patterns. These models are well-suited to capturing the nonlinear, strategic, and spatially correlated nature of poaching behavior. However, applying them to real-world poaching datasets presents two key challenges. First, the amount of historical data is typically small, especially in parks with limited monitoring capacity, whereas such generative models are usually data-hungry. Second, detection of poaching events is imperfect: many illegal activities go unnoticed during patrols, and negative labels indicating the absence of poaching are unreliable due to the difficulty of detecting well-hidden signs such as snares in dense vegetation~\citep{keane2011encounter}.

To address these challenges, we propose \ours, a framework based on \emph{latent composite flow matching}.
Building on occupancy models widely used in ecology~\citep{mackenzie2002estimating}, we represent poaching activity as a latent state, distinct from the probability of detection during patrols.
This separation enables explicit handling of imperfect detection: the latent state evolves through flow matching, while detection is modeled as the observation process in an occupancy framework.
To mitigate data scarcity, we further design a composite flow that initializes the generative process from a linear model’s prediction rather than random noise, which is standard in diffusion models.
By embedding domain knowledge in the initialization, our method improves generalization and supports effective learning in small-data settings.

We summarize our contributions are as follows: (1) We present \ours, the first application of generative AI to poaching prediction. (2) We propose a latent composite flow matching framework that jointly addresses the challenges of limited data and imperfect detection in conservation. (3) On real-world datasets from two national parks in Uganda, our method improves AUPR by 7.0\% and 10.2\% for MFNP and QENP, respectively. We also outline steps toward responsible deployment, including field pilots, and careful rollout.

\section{Related Works}

\paragraph{AI for poaching prediction.}

\citet{yang2014adaptive} introduced the Protection Assistant for Wildlife Security (PAWS) using a subjective-utility quantal response (SUQR) behavioral model. To explicitly account for imperfect detection, \citet{nguyen2016capture} proposed a two-layer Bayesian network with latent variables. However, \citet{kar2017cloudy} later found this approach brittle in practice due to model complexity; an ensemble of decision trees achieved higher accuracy with lower runtime and was validated in a one-month field test. To combine the strengths of prior methods, \citet{gholami2017taking} proposed a geo-clustering technique that yields a hybrid of multiple Markov random fields with a bagging ensemble of decision trees, supported by a five-month controlled field test. To further address nonuniform uncertainty stemming from uneven patrol effort, \citet{gholami2018adversary,xu2020stay} trained ensembles of weak learners—decision trees and Gaussian processes—on data stratified by patrol intensity, with  deployment via the SMART conservation platform~\citep{smart2013}. Finally, \citet{xu2021robust} used logistic regression to quantify the deterrence effects of ranger patrols on poaching risk.

Despite these practical successes, most approaches still rely on classical models with limited capacity to capture high-dimensional, nonlinear spatial patterns; few make use of modern deep learning. Most recently, \citet{kong2025robust} applied diffusion models to green security problems. Our work differs in three key ways:
(i) Their focus is on robust patrol optimization for general green security, whereas we focus specifically on poaching risk forecasting for conservation.
(ii) They do not account for imperfect detection, which is a critical challenge in conservation settings, while we explicitly model the visit-level detection process.
(iii) Their features are restricted to historical patrol effort, whereas we incorporate rich environmental covariates. In this richer feature space, diffusion models struggle due to their data-hungry nature. To address this, we propose a composite flow model that improves data efficiency.

\paragraph{Generative AI for wildlife conservation}
In camera‐trap vision, neural generative models have been applied directly to wildlife datasets: CycleGAN variants translate between sensor domains (e.g., visible and near-infrared) and augment rare species to improve few-shot classification~\citep{gao2020cyclegan, zhang2023few}. Beyond imagery, generative audio models have been developed to synthesize wildlife vocalizations and strengthen bioacoustic monitoring under data scarcity. Early work explored class-conditional GANs for animal-sound augmentation~\citep{kim2023dualdiscwavegan}, followed by diffusion-based pipelines that generate birdsong spectrograms to improve classifier accuracy~\citep{gibbons2024generative}, and more recently, diffusion models that synthesize anuran calls~\citep{manrique2025anuran}. Large language models (LLMs) have also been applied to detect illegal wildlife trafficking on online marketplaces by generating pseudo-labels for unlabeled listings~\citep{barbosa2025cost}. In contrast, our work focuses on generative modeling for poaching prediction in protected areas.

\section{Task, Challenges, and Preliminaries}

\subsection{Task Definition}

We study Murchison Falls National Park (MFNP, $\sim$5{,}000\,km$^2$) and Queen Elizabeth National Park (QENP, $\sim$2{,}500\,km$^2$) in Uganda. Both parks are critical for ecotourism and conservation and provide habitat for elephants, giraffes, hippos, and lions~\citep{critchlow2015spatiotemporal}.

We partition each park into a grid of $1\times 1$\,km cells, indexed by $i\in\{1,\dots,N\}$, and time into monthly intervals $t\in\{1,\dots,T\}$. For each  cell--month, we construct a feature vector $\mathbf{x}_{i,t}\in\mathbb{R}^d$ that includes \emph{static} geospatial attributes (aspect, elevation, surface water, seasonal water cover, and distances to rivers, patrol posts, and towns/villages) as well as \emph{dynamic} covariates (monthly average precipitation, temperature, and net primary productivity). The latent occupancy state, i.e., whether poaching activity occurs, is denoted by $z_{i,t}\in\{0,1\}$.

To combat poaching, rangers conduct GPS-tracked patrols, record observations, and remove wire snares that would otherwise trap endangered wildlife. Within month $t$, cell $i$ may be visited $J_{i,t}$ times. For visit $j\in\{1,\dots,J_{i,t}\}$, let $a_{i,t,j}\ge 0$ denote the \emph{visit-level patrol effort} (kilometers patrolled), and let $y_{i,t,j}\in\{0,1\}$ indicate whether illegal activity was detected during that visit.

\paragraph{Objective.}
Our goal is to model and forecast the true poaching risk, $p(z_{i,t}=1)$, using environmental covariates $\{\mathbf{x}_{i,t}\}$ and historical patrol effort $\{a_{i,t-1,j}\}$, while explicitly accounting for the current month’s visit-level effort $\{a_{i,t,j}\}$.

\begin{figure*}[t]
    \centering
    \includegraphics[width=0.85\linewidth]{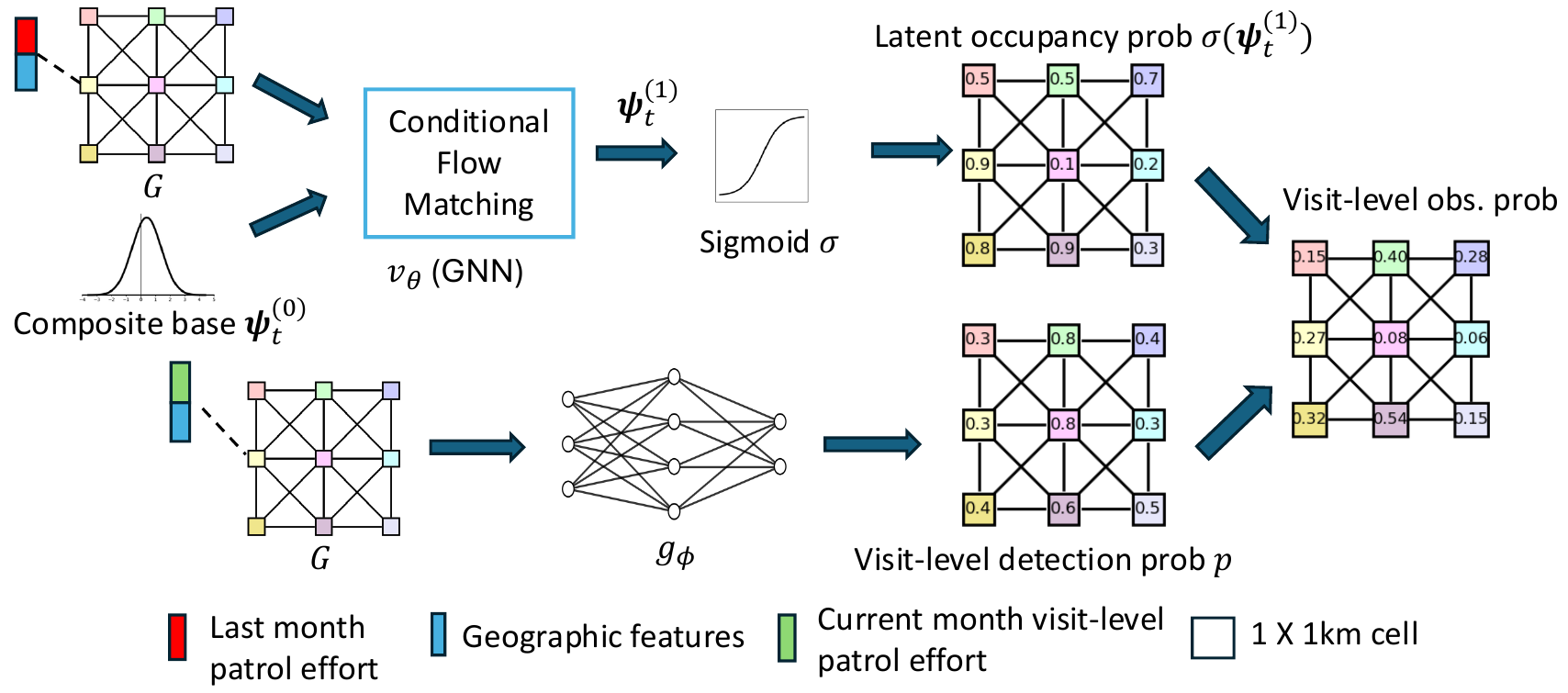}
\caption{Overview of the latent composite flow framework. \textbf{Upper branch (latent generation):} $1\times1$\,km cells are nodes in a graph $G$ with edges between adjacent cells. For month $t$, node features comprise geospatial covariates $\mathbf{x}_{i,t}$ and last month’s aggregated patrol effort $a^m_{i,t-1}
=\sum_j a_{i,t-1,j}$.
A \emph{composite base} initializes the latent logits $\boldsymbol{\psi}^{(0)}_t$ from a pretrained linear occupancy model with Gaussian perturbation. A graph-conditional velocity field $v_\theta$ transports $\boldsymbol{\psi}^{(0)}_t$ to $\boldsymbol{\psi}^{(1)}_t$ via conditional flow matching; $\sigma(\boldsymbol{\psi}^{(1)}_t)$ yields occupancy probabilities. \textbf{Lower branch (detection):} the visit-level detector $g_\phi(\mathbf{x}_{i,t},a_{i,t,j})$ uses geospatial features and current-month visit effort to produce detection probabilities $\{p_{i,t,j}\}$, which are combined with occupancy in the occupancy–detection likelihood.}

    \label{fig:overview}
\end{figure*}

\subsection{Challenges}

\paragraph{Complex poacher behavior}  
Poaching dynamics emerge from intertwined social and ecological drivers and adapt strategically to enforcement. Prior work highlights a deterrence–displacement effect: stronger patrol presence lowers activity within a cell but shifts risk to neighboring cells, leading to spatial spillovers~\citep{dancer2022evidence, rytwinski2024evidence, xu2021robust}. These dynamics yield non-stationary and spatially coupled patterns that simple discriminative models struggle to capture.

\paragraph{Imperfect detection}  
Observed incidents reflect \emph{detection}, not true occurrence. Because snares and traps are often well hidden, the likelihood of detection depends on factors such as habitat, visibility, and patrol effort~\citep{dobson2020making}. As a result, the data contain many false negatives and are biased by patrol routing. Models that overlook this distinction conflate absence with non-detection, leading to biased risk estimates. It is therefore essential to separate true poaching occurrence from the detection process.

\paragraph{Small data}
Although some parks have up to 20 years of patrol records, strong nonstationarity limits the \emph{effective} sample size: staffing levels, prices for illegal wildlife products
 and land–use change alter incentives and accessibility, making older data weak predictors of current risk. Following prior work~\citep{xu2020stay}, we therefore restrict training to the most recent $\leq$3 years to mitigate temporal drift. In addition, poaching patterns, ranger resources, and landscape features differ markedly across parks; naive cross‐park pooling induces severe dataset shift, so data cannot simply be combined to train a single universal model.

\subsection{Preliminaries on Flow Matching}
\label{sec:preliminaries}

To capture complex spatial structure, we adopt \emph{flow matching} (FM), a generative framework that learns a time–dependent velocity field to transport a base distribution toward the data distribution~\citep{lipman2023flow}. FM enables faster sampling than diffusion models~\citep{ho2020denoising}, while often achieving comparable or improved quality.

In this work, we focus on the \emph{conditional} version.  
At flow time $s{=}0$\footnote{Throughout, $t$ indexes months in the poaching data, while $s\in[0,1]$ parameterizes the generative flow.}, FM draws a base sample $\psi^{(0)}\!\sim\!p_0(\cdot\mid\mathbf{c})$ and evolves it under a conditional velocity field
$\frac{d\psi^{(s)}}{ds} \;=\; v_\theta\!\big(\psi^{(s)},\,s;\,\mathbf{c}\big),$
where $\mathbf{c}$ denotes contextual information. Integrating from $s{=}0$ to $s{=}1$ yields $\psi^{(1)}$, a sample from $p_{\text{data}}(\cdot\mid\mathbf{c})$. 

A key distinction from diffusion models is the choice of the initial distribution: FM permits $p_0$ to be any distribution, whereas diffusion models must begin with a standard Gaussian. Training FM commonly involves minimizing the loss
\begin{align}
\resizebox{\linewidth}{!}{$
\mathcal{L}_{\mathrm{FM}}(\theta)
= \mathbb{E}_{\substack{
   \mathbf{c}\sim p_{\mathrm{data}}(\mathbf{c}),\;
   \psi^{(1)}\sim p_{\mathrm{data}}(\cdot\mid \mathbf{c}),\\
   \psi^{(0)}\sim p_0(\cdot\mid \mathbf{c}),\;
   s\sim \mathcal{U}(0,1)}}
\Big\|\, v_\theta(\psi^{(s)}, s;\mathbf{c})
- \big(\psi^{(1)}-\psi^{(0)}\big)
\Big\|_2^2,
$}
\label{eq:flow}
\end{align}
where $\psi^{(s)} = (1-s)\psi^{(0)} + s\psi^{(1)}$. The linear interpolation ensures smooth trajectories, reducing ODE discretization error and improving sampling efficiency.  

While FM can capture complex spatial patterns, it does not by itself address (i) imperfect detection or (ii) small data.

\section{Proposed Method}

In this section, we propose \ours to address the challenges of imperfect detection and limited data.  

To handle imperfect detection, our framework builds on ecological occupancy models~\citep{mackenzie2002estimating}, which treat the true presence of poaching as a hidden state and explicitly model the probability of detection during patrols. Leveraging flow matching’s ability to capture complex, high-dimensional spatial patterns, we model the generative process of the latent occupancy state (i.e., the true presence of poaching) and add an explicit visit-level detection model on top. This design allows us to distinguish true absence from non-detection, thereby reducing dataset bias.

To address the limited-data challenge, we warm-start flow matching from a composite base initialized by a simple linear model. This initialization embeds domain knowledge, improves generalization, and enables more data-efficient learning than starting from random noise.

\subsection{Data Generation Process}
\label{sec:generation}

Figure~\ref{fig:overview} illustrates the data generation process of our framework. 
The design has two components: the \emph{upper branch}, which generates the probability of the latent occupancy states (i.e., the true but unobserved presence of poaching), and the \emph{lower branch}, which models the detection process during ranger patrols.  

\paragraph{Flow matching for latent occupancy generation.}
We represent each park as a grid graph $G=(V,E)$ of $1{\times}1$\,km cells, where nodes correspond to cells and edges connect neighbors. 
For month $t$, each cell $i\in V$ has features $\mathbf{c}_{i,t} = [\mathbf{x}_{i,t};\, a^m_{i,t-1}]$, combining geographic covariates $\mathbf{x}_{i,t}$ and the previous month’s aggregated patrol effort $a^m_{i,t-1} = \sum_{j} a^m_{i, t-1,j}$. 
This lagged effort captures patrol deterrence effects.

We place a conditional flow prior on the \emph{joint} vector of occupancy logits
$\boldsymbol{\psi}^{(1)}_{t}\in\mathbb{R}^{N}$ ($N{=}|V|$) and evolve it via
\[
\frac{d\boldsymbol{\psi}^{(s)}_t}{ds}
= v_\theta\!\big(\boldsymbol{\psi},\, s;\, G,\, \mathbf{C}_t\big),\quad
\boldsymbol{\psi}^{(0)}_t \sim p_0(\cdot \mid \mathbf{C}_t),\ \ s\in[0,1],
\]
where $\mathbf{C}_t$ is the collection of $\mathbf{c}_{i,t}$ over all nodes and $v_\theta$ is a message-passing graph neural network (GNN)~\citep{kipf2017semisupervised}.
Message passing propagates information across neighboring cells during the flow, enabling the model to capture spatial spillovers and local correlations in poaching risk. The final logits are mapped through a sigmoid to yield latent occupancy probabilities $r_{i,t} = \sigma(\psi_{i,t}^{(1)})$. 

\paragraph{Warm-starting from a composite base distribution.}
Instead of initializing the flow from uninformative Gaussian noise, we warm-start from a \emph{composite base}:  
\[
\boldsymbol{\psi}^{(0)}_{t} = b_\eta(\mathbf{C}_t) + \boldsymbol{\epsilon}, 
\qquad \boldsymbol{\epsilon}\sim \mathcal{N}(\mathbf{0},\sigma_0^2\mathbf{I}),
\]
where $b_\eta$ is a lightweight pretrained linear occupancy predictor and $\mathbf{C}_t$ collects node features.  
This base injects domain knowledge from simple ecological models while still allowing variability through the Gaussian perturbation, improving data efficiency in small-sample settings.

\paragraph{Visit-level detection model.}
The lower branch of Fig.~\ref{fig:overview} describes how rangers detect snares during patrols.  
Within each month $t$, a cell $i$ may be visited several times, where each visit $j$ involves patrol effort $a_{i,t,j}$ and results in a binary outcome $y_{i,t,j}\in\{0,1\}$ indicating whether snares were found.  
We model the detection probability with a lightweight neural head $g_\phi$ that takes environmental features and patrol effort as inputs:
$p_{i,t,j} = \sigma\!\big(g_\phi(\mathbf{x}_{i,t}, a_{i,t,j})\big).$
Since detections are only possible when the site is truly occupied, the model naturally separates \emph{true absence} from \emph{non-detection}.  
This design reduces bias in the observations and is consistent with ecological occupancy modeling practices.

Let $S_{i,t}=\mathbf{1}\{\exists j:\, y_{i,t,j}=1\}$ indicate that \emph{at least one} detection occurred in cell $i$ during month $t$.  Marginalizing $z_{i,t}$ gives the visit-level occupancy–detection log-likelihood:
\begin{align}
\mathcal{L}_{\text{occ}}(i,t)
&= \log\!\Big[ (1-r_{i,t})(1-S_{i,t}) \notag\\
&\qquad\quad +\; r_{i,t}\!
\prod_{j=1}^{J_{i,t}}
p_{i,t,j}^{\,y_{i,t,j}}\big(1-p_{i,t,j}\big)^{1-y_{i,t,j}}
\Big].
\label{eq:occ-likelihood}
\end{align}
We provide the full derivation of $\mathcal{L}_{\text{occ}}$ in  Appendix~\ref{app:derivation-occ-like}.

\subsection{Training and Inference Algorithm}
\label{sec:training}

As discussed in Section~\ref{sec:preliminaries}, flow matching (FM) requires samples from the \emph{target} distribution.  
Since the true latent logits $\boldsymbol{\psi}^{(1)}_t$ are unobserved, direct training is infeasible.  
Inspired by stable diffusion~\citep{rombach2022high}, we adopt a two-stage procedure.

\paragraph{Stage 1: Encoder--detector training.}  
In the first stage, we jointly train two components:  
(i) an encoder $f_\omega$ that maps monthly observations, environmental features and current month  patrol effort into latent occupancy logits $\hat{\boldsymbol{\psi}}_t$, and  
(ii) a detection head $g_\phi$ that predicts visit-level detection probabilities.  
The parameters $(\omega,\phi)$ are optimized by maximizing the summed occupancy--detection likelihood $\sum_{i,t}\mathcal{L}_{\text{occ}}(i,t)$.  
The resulting encoder produces surrogate logits $\hat{\boldsymbol{\psi}}_t^{(1)}$, which serve as training targets for the flow model in Stage~2.

\paragraph{Stage 2: Latent flow training.}  
In the second stage, we freeze the encoder--detector pair $(\omega,\phi)$ and train a conditional flow model $v_\theta$ to transport samples from the composite base $\boldsymbol{\psi}_t^{(0)}$ toward the surrogate logits $\hat{\boldsymbol{\psi}}_t^{(1)}$.

Following the linear--path flow matching objective in Eq.~\eqref{eq:flow}, we sample $s\!\sim\!\mathcal{U}[0,1]$ and form an interpolated state: $\boldsymbol{\psi}_t^{(s)} \;=\; (1-s)\,\boldsymbol{\psi}_t^{(0)} + s\,\hat{\boldsymbol{\psi}}_t^{(1)}.$
The velocity field $v_{\theta}$ is then trained to predict the straight-line displacement:
$\big\|\, v_\theta(\boldsymbol{\psi}_t^{(s)},\,s;\,G,\mathbf{C}_t)
   - (\hat{\boldsymbol{\psi}}_t^{(1)} - \boldsymbol{\psi}_t^{(0)}) \,\big\|_2^2.$
Here $(G,\mathbf{C}_t)$ provide conditioning context from the graph and features.

\paragraph{Inference.}  
For a forecast month $t^\star$, we construct features $\mathbf{C}_{t^\star}$ using covariates and past patrol effort.  
Sampling from the base distribution $\boldsymbol{\psi}^{(0)}_{t^\star}$ and integrating the velocity field yields latent logits $\boldsymbol{\psi}^{(1)}_{t^\star}$, converted to risk via $\sigma(\cdot)$.  Given planned patrol efforts $\{a_{i,t^\star,j}\}$, detection probabilities follow $p_{i,t^\star,j} = \sigma\!\big(g_\phi(\mathbf{x}_{i,t^\star}, a_{i,t^\star,j})\big)$. Note that the encoder is only used during training and not required at inference.

\begin{table*}[t]
\centering
\scriptsize
\setlength{\tabcolsep}{3pt}
\resizebox{\linewidth}{!}{
\begin{tabular}{llccccccc}
\toprule
Park & Year & LogReg & GP & MLP & GNN & Transformer & Diffusion & Ours \\
\midrule
\multirow{6}{*}{MFNP}
& 2017 & $0.305$ & $0.278$ \std{0.001} & $0.288$ \std{0.017} & $0.336$ \std{0.004} & $0.314$ \std{0.016} & $0.313$ \std{0.021} & \colorbox{green!25}{$\mathbf{0.374}$ \std{0.032}} \\
& 2018 & $0.344$ & $0.308$ \std{0.002} & $0.325$ \std{0.004} & $0.260$ \std{0.000} & $0.302$ \std{0.064} & $0.279$ \std{0.001} & \colorbox{green!25}{$\mathbf{0.377}$ \std{0.011}} \\
& 2019 & $0.406$ & $0.359$ \std{0.002} & $0.388$ \std{0.007} & \colorbox{green!25}{$\mathbf{0.526}$ \std{0.015}} & $0.475$ \std{0.141} & $0.362$ \std{0.003} & $0.409$ \std{0.009} \\
& 2020 & $0.421$ & $0.401$ \std{0.003} & $0.408$ \std{0.004} & $0.366$ \std{0.014} & $0.327$ \std{0.002} & $0.439$ \std{0.023} & \colorbox{green!25}{$\mathbf{0.473}$ \std{0.006}} \\
& 2021 & $0.360$ & $0.364$ \std{0.011} & $0.335$ \std{0.010} & \colorbox{yellow!25}{$\mathbf{0.430}$ \std{0.023}} & $0.295$ \std{0.085} & $0.342$ \std{0.032} & \colorbox{yellow!25}{$\mathbf{0.423}$ \std{0.004}} \\
\rowcolor{gray!12}
& Avg  & $0.367$ & $0.342$ & $0.349$ & $0.384$ & $0.343$ & $0.347$ & $\mathbf{0.411}$ \\
\midrule
\multirow{4}{*}{QENP}
& 2014 & $0.119$ & $0.107$ \std{0.003} & $0.103$ \std{0.003} & \colorbox{green!25}{$\mathbf{0.136}$ \std{0.013}} & $0.100$ \std{0.002} & $0.102$ \std{0.003} & $0.116$ \std{0.009} \\
& 2015 & $0.180$ & $0.156$ \std{0.002} & $0.172$ \std{0.006} & $0.111$ \std{0.000} & $0.173$ \std{0.016} & $0.162$ \std{0.028} & \colorbox{green!25}{$\mathbf{0.201}$ \std{0.017}} \\
& 2016 & $0.258$ & $0.220$ \std{0.005} & $0.227$ \std{0.004} & $0.238$ \std{0.009} & $0.220$ \std{0.019} & $0.201$ \std{0.034} & \colorbox{green!25}{$\mathbf{0.299}$ \std{0.027}} \\
\rowcolor{gray!12}
& Avg  & $0.186$ & $0.161$ & $0.167$ & $0.162$ & $0.164$ & $0.155$ & $\mathbf{0.205}$ \\
\bottomrule
\end{tabular}}
\caption{AUPR comparison on MFNP and QENP. All methods use a linear detection head for a fair comparison. Cells in \colorbox{green!25}{green} indicate a clear winner (best mean and the runner-up lies outside the winner’s std), while \colorbox{yellow!25}{yellow} denotes a practical tie (within one std of the best). LogReg is trained with deterministic optimization and therefore has no variance.}
\label{table:main}
\end{table*}

\section{Experiments}

\subsection{Experimental Setup}

\paragraph{Data processing}  
We use ranger patrol and poaching records from Murchison Falls National Park (MFNP, 2014–2021) and Queen Elizabeth National Park (QENP, 2011–2016), provided by the Uganda Wildlife Authority. Following \citep{kong2025robust}, we focus on historically high-risk subregions by constructing groups of cells each month: we first select the 20 cells with the highest observed poaching counts, then iteratively expand each group by adding adjacent cells until reaching a maximum of 25 cells. To address non-stationarity, we adopt the windowing strategy of \citep{xu2020stay}, where for a given test year, only the previous three years of data are used for training.

\paragraph{Baselines}

For a fair comparison, we use the same linear detection head across all methods, varying only the model used to predict the latent occupancy state. Each baseline is trained by maximizing the log-likelihood of visit-level observations. Specifically, we compare the following models:

\begin{itemize}
    \item \textbf{Logistic Regression (LogReg).} A linear occupancy model optimized jointly with the detection head using the BFGS optimizer~\citep{nocedal2006numerical}.  
    \item \textbf{Gaussian Process (GP).} Trained with the detection head using variational inference with inducing points. The objective combines the expected log-likelihood of observed snares with a KL divergence regularizer toward the prior.  
    \item \textbf{Multilayer Perceptron (MLP).} A three-layer neural network trained with the AdamW optimizer~\citep{loshchilov2019decoupled}.  
    \item \textbf{Graph Neural Network (GNN).} Encodes spatial relationships among cells via graph convolutional layers~\citep{kipf2017semisupervised}. 
    \item \textbf{Transformer.} Uses self-attention layers to propagate information across cells, capturing long-range dependencies and assigning adaptive weights to neighbors, unlike GNNs which aggregate uniformly ~\citep{loshchilov2019decoupled}.  
    \item \textbf{Diffusion Model.} Trained in the same two-stage framework as our method, but replacing composite flow matching with a conditional denoising diffusion model.  
\end{itemize}

We provide further experimental details in Appendix~\ref{appendix:exp-details}.

\paragraph{Metric}  
Since the latent occupancy state is unobserved, we evaluate performance at the observation level.  
This aligns with the practical goal: given a fixed patrol effort, the key question is whether rangers can detect any poaching activity in a cell during a given month.  
We therefore compute the probability of at least one detection as
$\hat{p}_{\text{any}, i,t}
= r_{i,t}\!\left(1 - \prod_{j=1}^{J_{i,t}} (1 - p_{i,t,j})\right),$
and compare it against the binary monthly label $y_{i,t}$, indicating whether any poaching was observed in that month.  We then 
evaluate predictions using the area under the precision--recall curve (AUPR), which is particularly appropriate for this task given the severe class imbalance.

\subsection{Main Results}

Table~\ref{table:main} reports results on MFNP and QENP. We summarize our findings as follows.  

(1) Our method significantly outperforms baselines on both datasets. On MFNP, we achieve clear improvements in 3 out of 5 years and match the strongest baseline in 1 year. On QENP, we achieve 2 clear wins out of 3 years. On average, \ours improves AUPR by $7.0\%$ on MFNP and $10.2\%$ on QENP compared to the strongest baseline.  (2) Surprisingly, the linear occupancy model (LogReg) often performs competitively, and in some cases even outperforms deep learning methods. We attribute this to the small and noisy nature of the data, where simpler models are less prone to overfitting. This highlights the need for caution and careful design when applying deep learning models in this domain. (3) Diffusion models do not perform well in our setting and we found two main reasons.  
(i) Training is less stable than flow matching, likely due to the stochastic sampling process.  
(ii) When complex environmental features are included, diffusion model performance gets stuck, highlighting the data-hungry nature of diffusion models. In contrast, our method solves this problem with a more data-efficient composite base distribution.

\subsection{Ablation Study}

\begin{table}[t]
\centering
\small
\setlength{\tabcolsep}{4pt}
\begin{tabular}{cccc}
\toprule
Year & w/o composite base & w/o detection head & Ours \\
\midrule
2017 & $0.341$ \std{0.008} & $0.278$ \std{0.011} & $\mathbf{0.374}$ \std{0.032} \\
2018 & $0.368$ \std{0.019} & $0.265$ \std{0.006} & $\mathbf{0.377}$ \std{0.011} \\
2019 & $0.382$ \std{0.006} & $0.301$ \std{0.006} & $\mathbf{0.409}$ \std{0.009} \\
2020 & $0.429$ \std{0.007} & $0.350$ \std{0.005} & $\mathbf{0.473}$ \std{0.006} \\
2021 & $0.338$ \std{0.040} & $0.290$ \std{0.009} & $\mathbf{0.423}$ \std{0.004} \\
\bottomrule
\end{tabular}
\caption{Ablation study on MFNP, evaluated using AUPR.}
\label{table:ablation}
\end{table}

We evaluate the contribution of each model component through ablation studies. As shown in Table~\ref{table:ablation}, removing either the composite base or the detection head leads to a consistent drop in AUPR across all years. These results demonstrate that both components are essential, and their combination yields the best overall performance.

\subsection{Case Study}
In this subsection, we conduct a case study to examine where \ours works particularly well.  
For each cell $i$ at time $t$, we compute the log loss $l_{i,t}^{\text{model}} = - y_{i,t}\,\log(\hat{p}_{\text{any}, i,t}) - (1-y_{i,t})\,\log(1-\hat{p}_{\text{any}, i,t})$.
Log loss is a proper scoring rule~\citep{gneiting2007strictly}, which evaluates both accuracy and calibration of probabilistic forecasts.

To assess relative performance, we calculate the log-loss difference relative to LogReg: $l_{i,t}^{\text{diff}} = l_{i,t}^{\text{\ours}} - l_{i,t}^{\text{LogReg}}.$
Smaller values of $l_{i,t}^{\text{diff}}$ indicate that our method achieves lower loss and thus more accurate forecasts. We then compare the environmental and patrol-effort features of cells in the 10th percentile of $l_{i,t}^{\text{diff}}$ (i.e., where our approach most strongly outperforms LogReg) against the feature distribution of all cells in the same test year.

\begin{figure}[t]
    \centering
    \includegraphics[width=\linewidth]{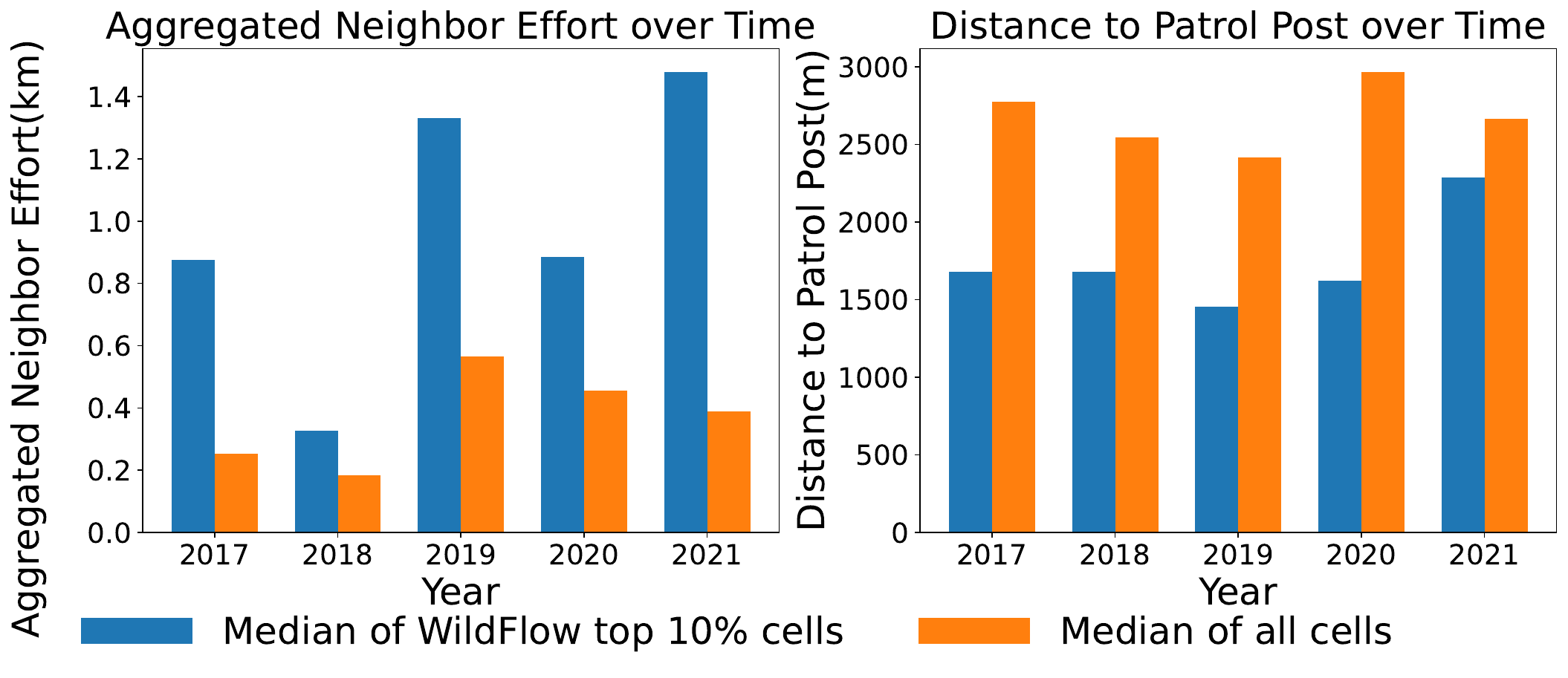}
    \caption{Case study of two predictive features on MFNP.
We compare the median values of each feature across all cells with those from the subset of top 10\% cells where WildFlow achieves the largest relative improvement over logistic regression.}
\label{fig:case_study}
\end{figure}

In Figure~\ref{fig:case_study}, we observe that \ours consistently achieves stronger performance on cells with higher levels of adjacent patrol effort across all test years. Neighboring patrol effort can induce displacement effects and create intricate spatial dependencies,\footnote{We intentionally do not display spatial maps of poaching predictions in the paper for security reasons.} and the superior performance of our method in these regions supports the intuition that flow matching is well-suited to modeling complex, high-dimensional spatial patterns. We also find that \ours tends to perform better on cells located closer to patrol posts. A possible explanation is that proximity to patrol posts increases both the frequency and reliability of detections, which our method is better able to exploit. Exploring the interpretability of these spatial patterns remains an interesting direction for future work.

\section{Path to Deployment}
Our objective is to translate the proposed latent composite flow framework into a field-ready tool that guides patrol allocation under real operational constraints (limited compute, bandwidth, and staff time). We proceed in two stages:  (1) \emph{validate utility} via prospective field pilots, and (2) \emph{deploy at scale} with ongoing monitoring.

\paragraph{Prospective field pilots.} We will run pilots in MFNP and Cross River National Park, in collaboration with the Uganda Wildlife Authority and the Nigeria National Park Service.
 Following \citet{xu2020stay}, we will focus on regions that had low patrol effort in the past to reduce bias from prior ranger knowledge. The model gives each cell a risk score, which we place into three groups: high, medium, and low. Rangers receive patrol plans without seeing these risk labels, and we keep patrol effort similar across the three groups. We will measure effectiveness by (i) snares found per km patrolled and (ii) whether higher-risk groups yield more detections than lower-risk groups .

\paragraph{Responsible rollout.}
Contingent on conclusive field-trial results, we will integrate the tool with SMART conservation software~\citep{smart2013} for broader use, conduct ranger workshops to incorporate practitioner feedback, and compare model guidance with ranger intuition. Deployment will be staged (pilot $\rightarrow$ phased scale-up) with a continuous feedback loop to audit and mitigate bias, and to update models on a regular cadence.

\section{Conclusion}
In this paper, we present \ours, the first generative AI-based approach for poaching prediction that explicitly addresses practical challenges such as observation bias and data scarcity. \ours achieves substantial improvements over various baselines on datasets from two national parks in Uganda.
These results establish a strong foundation for applying generative AI techniques in wildlife conservation. As next steps, we aim to further reduce inference costs and conduct rigorous field evaluations to assess the framework’s real-world effectiveness, with the ultimate goal of enabling large-scale deployment.

\bibliography{ref}

\newpage
\appendix
\section{Appendix}
\subsection{Derivation of the Marginalized Log-Likelihood}
\label{app:derivation-occ-like}

For cell--month $(i,t)$, let $r_{i,t}=\sigma(\psi_{i,t})$ and 
$\mathbf y_{i,t}=(y_{i,t,1},\ldots,y_{i,t,J_{i,t}})$ with detection
probabilities $\{p_{i,t,j}\}_{j=1}^{J_{i,t}}$. Assume:
(i) $z_{i,t}\sim\mathrm{Bernoulli}(r_{i,t})$;
(ii) visits are conditionally independent given $z_{i,t}$;
(iii) no false positives: $P(y_{i,t,j}=1 \mid z_{i,t}=0)=0$.

We marginalize over $z_{i,t}\in\{0,1\}$:
\begin{align}
P(\mathbf y_{i,t}\,|\,\psi_{i,t},\{p_{i,t,j}\})
&= (1-r_{i,t})\,P(\mathbf y_{i,t}|z{=}0) \notag\\
&\quad + r_{i,t}\,P(\mathbf y_{i,t}|z{=}1).
\label{eq:marg}
\end{align}

By (iii),
\[
P(\mathbf y_{i,t}|z{=}0)=\mathbf 1\{\forall j: y_{i,t,j}=0\},
\]
and by (ii),
\[
P(\mathbf y_{i,t}|z{=}1)
= \prod_{j=1}^{J_{i,t}} p_{i,t,j}^{\,y_{i,t,j}}\,(1-p_{i,t,j})^{1-y_{i,t,j}}.
\]

Substituting into \eqref{eq:marg} yields
\begin{align}
P(\mathbf y_{i,t}\,|\,\psi_{i,t},\{p_{i,t,j}\})
&= (1-r_{i,t})\,\mathbf 1\{\forall j: y_{i,t,j}=0\} \notag\\
&\quad + r_{i,t}\!\prod_{j=1}^{J_{i,t}} 
      p_{i,t,j}^{\,y_{i,t,j}}(1-p_{i,t,j})^{1-y_{i,t,j}}.
\label{eq:lik_full}
\end{align}

Define
\[
S_{i,t}=\mathbf 1\{\exists j: y_{i,t,j}=1\},\quad
L_{i,t}=\prod_{j=1}^{J_{i,t}} p_{i,t,j}^{\,y_{i,t,j}} (1-p_{i,t,j})^{1-y_{i,t,j}}.
\]
Since $\mathbf 1\{\forall j: y_{i,t,j}=0\}=1-S_{i,t}$, the log-likelihood term is
\[
\log P(\mathbf y_{i,t}|\cdot)
= \log\!\big((1-r_{i,t})(1-S_{i,t}) + r_{i,t}\,L_{i,t}\big),
\]
and the training loss is its negative.

\textbf{Special cases.}
If $S_{i,t}=0$ (no detections),
\[
\log P(\mathbf y_{i,t}|\cdot)
= \log\!\big((1-r_{i,t}) + r_{i,t}\prod_{j}(1-p_{i,t,j})\big).
\]
If $S_{i,t}=1$ (at least one detection),
\begin{align}
\log P(\mathbf y_{i,t}|\cdot) 
&= \log r_{i,t} \notag\\
&\quad + \sum_{j}\Big(
      y_{i,t,j}\log p_{i,t,j} \notag\\
&\qquad\quad + (1-y_{i,t,j})\log(1-p_{i,t,j})
    \Big).
\end{align}

\subsection{Training Algorithm}
\label{app:training}

We provide the full training algorithm in Algorithm~\ref{alg:two-stage}.
 \paragraph{Stage 1: Training encoder and detector.}  
In Stage~1, we jointly train two components:  
(i) an encoder that estimates latent occupancy logits, and  
(ii) a detection head that models visit-level detection probabilities.  

For each cell $i$ at month $t$, the node feature is defined as  
\[
\mathbf{c}'_{i,t} = [\mathbf{x}_{i,t},\, S_{i,t},\, a^m_{i,t}],
\]  
where $\mathbf{x}_{i,t}$ are geographic covariates, $S_{i,t} = \mathbf{1}\{\exists j:\, y_{i,t,j}=1\}$ indicates whether poaching was observed, and $a^m_{i,t} = \sum_j a_{i,t,j}$ is the total patrol effort that month. Let $\mathbf{C}'_t$ denote the collection of all node features in month $t$.  

The encoder maps the graph and node features to latent occupancy logits:  
\[
\hat{\boldsymbol{\psi}}^{(1)}_{t} = f_\omega(G, \mathbf{C}'_t).
\]  

Independently, the detection head computes the probability of detecting a snare on visit $j$ as  
\[
p_{i,t,j} = \sigma\!\big(g_\phi(\mathbf{x}_{i,t}, a_{i,t,j})\big),
\]  
which depends only on visit-level covariates and patrol effort.  

Substituting $r_{i,t}=\{\sigma(\hat{\psi}_{i,t})\}$ and $\{p_{i,t,j}\}$ into Eq.~\eqref{eq:occ-likelihood} yields the occupancy–detection log-likelihood $\mathcal{L}_{\text{occ}}(i,t)$. We then jointly optimize encoder and detector parameters $(\omega,\phi)$ by solving  
\[
\max_{\omega,\phi}\; \sum_{i,t} \mathcal{L}_{\text{occ}}(i,t).
\]

\paragraph{Stage 2: Training latent flow matching.}  
In Stage~2, we train a conditional flow prior to transport a composite base distribution to the latent targets inferred by the Stage~1 encoder. At this stage, the encoder and detector parameters $(\omega,\phi)$ are frozen, and the encoder outputs $\hat{\boldsymbol{\psi}}^{(1)}_{t}$ serve as the training targets.  

We train a graph–conditional velocity field $v_\theta(\cdot,\cdot;\,G,\mathbf{C}_t)$ using the composite base
$\boldsymbol{\psi}^{(0)}_{t}=b_\eta(\mathbf{C}_t)+\boldsymbol{\epsilon}$ with
$\boldsymbol{\epsilon}\!\sim\!\mathcal{N}(\mathbf{0},\sigma_0^2\mathbf{I})$,
and straight-line paths
$\boldsymbol{\psi}^{(s)}_{t}=(1-s)\boldsymbol{\psi}^{(0)}_{t}+s\,\hat{\boldsymbol{\psi}}^{(1)}_{t}$, $s\!\in\![0,1]$. Prior work has shown that when the base distribution is closer to the target distribution, the generalization performance of flow matching improves~\citep{kong2025composite}.

Putting the components together, the conditional flow-matching objective is defined as
\begin{equation}
\resizebox{\linewidth}{!}{$
\min_{\theta}\;\sum_{t}\;
\mathbb{E}_{\substack{s\sim \mathcal{U}[0,1]\\ \boldsymbol{\epsilon}\sim \mathcal{N}(\mathbf{0},\sigma_0^2\mathbf{I})}}
\Big\|
v_\theta\!\big(\boldsymbol{\psi}^{(s)}_{t},\, s;\, G, \mathbf{C}_t\big)
-\big(\hat{\boldsymbol{\psi}}^{(1)}_{t}-\boldsymbol{\psi}^{(0)}_{t}\big)
\Big\|_2^2
$}
\end{equation}

\paragraph{Inference.}  
For a forecast month $t^\star$, we construct $\mathbf{C}_{t^\star}$ from features and \emph{past} patrol effort only. We then sample  an initial base  
$\boldsymbol{\psi}^{(0)}_{t^\star} \sim p_0(\cdot \mid \mathbf{C}_{t^\star})$,  
and integrate the learned velocity field  
\[
\frac{d\boldsymbol{\psi}^{(s)}_{t^\star}}{ds}
= v_\theta\!\big(\boldsymbol{\psi}^{(s)}_{t^\star}, s;\, G, \mathbf{C}_{t^\star}\big),
\quad s \in [0,1],
\]  
to obtain $\boldsymbol{\psi}^{(1)}_{t^\star}$. The resulting occupancy risk is  
$\mathbf{r}_{t^\star} = \sigma(\tilde{\boldsymbol{\psi}}_{t^\star})$. For more stable estimates, we can draw $M$ samples and compute the Monte Carlo mean.

Given planned visit-level efforts $\{a_{i,t^\star,j}\}_{j=1}^{J_{i,t^\star}}$, detection probabilities are obtained from the trained head:  
\[
p_{i,t^\star,j} = \sigma\!\big(g_\phi(\mathbf{x}_{i,t^\star}, a_{i,t^\star,j})\big).
\]  
The probability of \emph{at least one} detection in cell $i$ during month $t^\star$ is then  
\begin{align}
\hat{p}_{\text{any}, i,t^\star}
= r_{i,t^\star}\Big(1 - \prod_{j=1}^{J_{i,t^\star}} (1 - p_{i,t^\star,j})\Big).
\label{eq:any}
\end{align}
Note that the encoder is only used during training and is not required at inference time.

\begin{algorithm}[!t]
\small
\DontPrintSemicolon
\caption{Two-Stage Training for \ours}
\label{alg:two-stage}
\KwInput{
Monthly graphs $\{G_t=(V,E)\}$ with node features $\mathbf{C}_t$ and $\mathbf{C}'_t$; 
visit-level data $\{(a_{i,t,j},\,y_{i,t,j})\}$; 
encoder $f_\omega$, detection head $g_\phi$, flow velocity $v_\theta$; 
pretrained linear occupancy model $b_{\eta}$.
}
\KwOutput{
Trained $f_\omega, g_\phi, v_\theta$.
}

\BlankLine
\textbf{Stage 1:\; Encoder \& Detection Head}\;

\ForEach{epoch $=1,\dots,E_1$}{
  \ForEach{month $t$}{
    $\hat{\boldsymbol{\psi}}_t \leftarrow f_\omega(G_t,\,\mathbf{C}'_t)$ 
    \Comment*[r]{logits for latent occupancy}
    \ForEach{cell $i\in V$ and visit $j=1,\dots,J_{i,t}$}{
      $p_{i,t,j} \leftarrow \sigma\!\big(g_\phi(\mathbf{x}_{i,t},\,a_{i,t,j})\big)$ 
      \Comment*[r]{visit-level detection prob.}
    }
    $\mathcal{L}_{\text{occ}}(t) \leftarrow \sum_{i}\log p\!\big(\{y_{i,t,j}\}_j \mid \hat{\psi}_{i,t}, \{p_{i,t,j}\}_j\big)$ 
    \Comment*[r]{Eq.~\eqref{eq:occ-likelihood}}
  }
  Update $(\omega,\phi)$ by \emph{maximizing} $\sum_t \mathcal{L}_{\text{occ}}(t)$ 
}
\Comment*[l]{Outputs surrogate targets $\hat{\boldsymbol{\psi}}^{(1)}_t \gets f_\omega(G_t,\mathbf{C}'_t)$.}

\BlankLine
\textbf{Stage 2:\; Latent Flow Matching}\;

Freeze $(\omega,\phi)$\;
\ForEach{epoch $=1,\dots,E_2$}{
  \ForEach{month $t$}{
    Sample base $\boldsymbol{\psi}^{(0)}_t \sim \mathcal{N}\!\big(b_\eta(\mathbf{C}_t),\,\sigma_0^2 I\big)$\;
    Draw $s\sim \mathcal{U}[0,1]$ and form $\boldsymbol{\psi}^{(s)}_t=(1-s)\boldsymbol{\psi}^{(0)}_t+s\,\hat{\boldsymbol{\psi}}^{(1)}_t$\;
    Compute target velocity $u_t=\hat{\boldsymbol{\psi}}^{(1)}_t-\boldsymbol{\psi}^{(0)}_t$\;
    Predict $v_t = v_\theta\!\big(\boldsymbol{\psi}^{(s)}_t,\,s;\,G_t,\mathbf{C}_t\big)$\;
    $\mathcal{L}_{\text{FM}}(t) \leftarrow \|\,v_t - u_t\,\|_2^2$\;
  }
  Update $\theta$ by \emph{minimizing} $\sum_t \mathcal{L}_{\text{FM}}(t)$
}
\Return $(f_\omega, g_\phi, v_\theta)$
\end{algorithm}

\subsection{Experimental Details}
\label{appendix:exp-details}

\paragraph{LogReg:} 
The occupancy model is linear, and the detection component is modeled with a linear head, consistent with the other baselines. Predictions are made independently for each cell. To capture spatial spillovers from patrol activity, we follow \citet{xu2021robust} and include aggregated  patrol effort in adjacent cells as an additional feature in the occupancy model.  
We jointly train the linear occupancy model and detection head by minimizing  
$-\sum_{i,t}
\mathcal{L}_{\text{occ}}(i,t),$ 
where $\mathcal{B}$ denotes the batch. Optimization is performed using the BFGS algorithm from SciPy~\citep{2020SciPy-NMeth}.

\noindent\textbf{MLP:} 
We implement a multilayer perceptron (MLP) baseline for occupancy modeling, using a three-layer fully connected network with ReLU activations and dropout regularization. To ensure numerical stability, the output logits are clipped to an absolute value of 10 before applying the sigmoid transformation. Predictions are made independently for each cell. To capture spatial spillovers from patrol activity, we follow \citet{xu2021robust} and include aggregated  patrol effort in adjacent cells as an additional feature in the occupancy model.
The model is trained by minimizing $-\sum_{i,t}
\mathcal{L}_{\text{occ}}(i,t)$. The MLP occupancy network and detection head are optimized jointly using the AdamW optimizer, with hyperparameters listed in Table~\ref{tab:hyperparams_all}.

\paragraph{GP:} 
We model occupancy using a sparse variational Gaussian process (GP), where the latent function $f$ is mapped to occupancy probability via a sigmoid transformation. Predictions are made independently for each cell. To capture spatial spillovers from patrol activity, we follow \citet{xu2021robust} and include aggregated  patrol effort in adjacent cells as an additional feature in the occupancy model. The GP and linear detection head are trained jointly by maximizing the evidence lower bound (ELBO):  
\[
\mathcal{L}_{\text{GP}} \;=\; 
-\,\mathbb{E}_{q(f)}\!\left[ \log p\big(y \mid f\big) \right] 
\;+\; \mathrm{KL}\!\left( q(f) \,\|\, p(f) \right),
\]  
where $p(f)$ is the GP prior and $q(f)$ is a sparse variational posterior defined using inducing points to approximate the true posterior. The expectation is estimated via Monte Carlo sampling. We use the GPyTorch implementation~\citep{gardner2018gpytorch}, and the detailed hyperparameter setup is provided in Table~\ref{tab:hyperparams_all}.

\paragraph{GNN:} 
We represent each cell as a node in a graph, with edges defined by spatial adjacency. Node features and graph structure are passed through a GCN encoder~\citep{kipf2017semisupervised}, which outputs an occupancy probability for each node via a sigmoid transformation. The encoder and detection head are trained jointly by minimizing $-\sum_{i,t} \mathcal{L}_{\text{occ}}(i,t)$. We optimize using AdamW~\citep{loshchilov2019decoupled} and apply gradient clipping with a maximum norm of 5. Hyperparameter details are provided in Table~\ref{tab:hyperparams_all}.

\paragraph{Transformer:} 
As with the GNN, we represent each cell as a node in a graph, with edges defined by spatial adjacency. Node features and graph structure are passed through a Transformer-based graph encoder, which performs multi-head attention over each node’s neighborhood and outputs an occupancy probability via a sigmoid transformation. The encoder and detection head are trained jointly by minimizing $-\sum_{i,t} \mathcal{L}_{\text{occ}}(i,t),$  
using AdamW~\citep{loshchilov2019decoupled} with gradient clipping at a maximum norm of 5. Hyperparameter details are provided in Table~\ref{tab:hyperparams_all}.

\paragraph{Diffusion Model:} 
We adopt the same two-stage training procedure as \ours. Hyperparameter settings are listed in Table~\ref{tab:hyperparams_all}.  

\paragraph{\ours:} 
The hyperparameter settings for our method are also summarized in Table~\ref{tab:hyperparams_all}.

\begin{table}[t]
\centering
\begin{tabular}{llp{3cm}}
\toprule
\textbf{Model} & \textbf{HP} & \textbf{Value} \\
\midrule
\multirow{6}{*}{MLP} 
 & Epochs & 100 \\
 & LR & $\{3, 5, 8\}\times 10^{-3}$ \\
 & Optimizer & AdamW \\
 & Batch & 512 \\
 & W. Decay & $10^{-4}$ \\
 & Logit Bound & 10 \\
\midrule
\multirow{5}{*}{GP}  
 & Epochs & 100 \\
 & LR & $\{3, 5\}\times 10^{-3}$ \\
 & Optimizer & AdamW \\
 & Batch & 256 \\
 & MC Samples & 8 / 50 \\
\midrule
\multirow{6}{*}{GNN} 
 & Epochs & 120 \\
 & LR & $\{3, 5\}\times 10^{-3}$ \\
 & Optimizer & AdamW \\
 & Batch & 256 \\
 & Hidden Dim & 128 \\
 & Layers & 2 \\
\midrule
\multirow{7}{*}{Transformer} 
 & Epochs & 120 \\
 & LR & $\{3, 5\}\times 10^{-3}$ \\
 & Optimizer & AdamW \\
 & Batch & 256 \\
 & Hidden Dim & 128 \\
 & Layers & 2 \\
 & Heads & 4 \\
\midrule
\multirow{5}{*}{Diffusion (Stage I)} 
 & Optimizer & AdamW \\
 & LR & $10^{-2}$ \\
 & Batch & 256 \\
 & Hidden Dim & 128 \\
 & Layers & 2 \\
\midrule
\multirow{5}{*}{Diffusion (Stage II)} 
 & Optimizer & AdamW \\
 & LR & $\{10^{-2}, 10^{-3}\}$ \\
 & Batch & 256 \\
 & Hidden Dim & 128 \\
 & Layers & 2 \\
\midrule
\multirow{5}{*}{\ours (Stage I)} 
 & Optimizer & AdamW \\
 & LR & $10^{-2}$ \\
 & Batch & 256 \\
 & Hidden Dim & 128 \\
 & Layers & 2 \\
\midrule
\multirow{6}{*}{\ours (Stage II)} 
 & Optimizer & AdamW \\
 & LR & $\{10^{-2}, 10^{-3}\}$ \\
 & Batch & 256 \\
 & Hidden Dim & 128 \\
 & Layers & 2 \\
 & $\sigma_0$ & 0.1 \\
\bottomrule
\end{tabular}
\caption{Hyperparameter settings for all methods.}
\label{tab:hyperparams_all}
\end{table}

\end{document}